\documentclass[letterpaper, 10 pt, journal, twoside]{IEEEtran}

\usepackage{xspace}
\usepackage[dvipsnames]{xcolor}
\usepackage{graphicx}
\usepackage{amsmath}
\usepackage{amssymb}
\usepackage{mathtools}
\usepackage{nicefrac}
\usepackage{bm}

\usepackage{adjustbox}
\usepackage{booktabs}
\usepackage{multirow}
\usepackage{graphicx}
\usepackage{tikz}
\usepackage{siunitx}
\usepackage{colortbl}
\usepackage{csquotes}
\usepackage{pifont}
\usepackage{xspace}
\usepackage{microtype}

\usetikzlibrary{positioning,arrows.meta,calc,fit,backgrounds}
\usetikzlibrary{decorations.pathreplacing,calligraphy}

\makeatletter
\let\NAT@parse\undefined
\makeatother

\usepackage[hidelinks]{hyperref}
\usepackage{cite}
\usepackage[capitalise]{cleveref}

\newcommand{\STQs}{STQ$_\text{1}$\xspace}
\newcommand{\AQs}{AQ$_\text{1}$\xspace}

\DeclareSIUnit{\pp}{\textup{p.p.}}

\let\tensor\bm

\makeatletter
\newcommand*{\@rowstyle}{}
\newcommand*{\rowstyle}[1]{%
  \gdef\@rowstyle{#1}%
  \@rowstyle\ignorespaces%
}
\newcolumntype{=}{>{\gdef\@rowstyle{}}}
\newcolumntype{+}{>{\@rowstyle}}
\makeatother

\newcommand{\grayrow}{\rowstyle{\color{gray}}}
\newcommand{\resetrow}{\rowstyle{\color{black}}}

\newcommand{\tabna}{\textbf{---}}

\newcommand{\tabbf}[1]{\textbf{\tablenum{#1}}}
\newcommand{\tabul}[1]{\underline{\tablenum{#1}}}

\newcommand{\cmark}{\ding{51}}%
\newcommand{\xmark}{\ding{55}}%

\newcommand{\tinycolorbox}[2]{\tikz[baseline=(a.base),inner sep=0pt]\node[fill=#1](a){#2\strut};}

\newcommand{\colorboxmark}[1]{%
  \begin{tikzpicture}[baseline={(current bounding box.south)}]
    \node[
      rectangle,
      rounded corners=0.5pt,
      fill=#1,
      draw=#1!50!black,
      line width=0.5pt,
      inner sep=0pt,
      minimum height=1.1ex,
      minimum width=1.1ex,
    ] {};
  \end{tikzpicture}%
}

\definecolor{Y1}{HTML}{FAF2D2}%
\definecolor{Y2}{HTML}{F4E8C1}%
\definecolor{Y3}{HTML}{EFE0AF}%
\definecolor{Y4}{HTML}{E9D79D}%
\definecolor{Y5}{HTML}{E3CD8C}%
\definecolor{Y6}{HTML}{DCC482}%

\definecolor{R1}{HTML}{FCE8E6}%
\definecolor{R2}{HTML}{F6D8D4}%
\definecolor{R3}{HTML}{F0C8C3}%
\definecolor{R4}{HTML}{EAB8B1}%
\definecolor{R5}{HTML}{DDA6A0}%
\definecolor{R6}{HTML}{D09890}%

\definecolor{P1}{HTML}{F3ECF7}%
\definecolor{P2}{HTML}{E5D7EB}%
\definecolor{P3}{HTML}{D8C3E0}%
\definecolor{P4}{HTML}{CDB5D6}%
\definecolor{P5}{HTML}{C3A8CD}%
\definecolor{P6}{HTML}{B79BC2}%

\definecolor{B1}{HTML}{EDF3FA}%
\definecolor{B2}{HTML}{DCE6F2}%
\definecolor{B3}{HTML}{C7D4E4}%
\definecolor{B4}{HTML}{B6C4D8}%
\definecolor{B5}{HTML}{A6B6CC}%
\definecolor{B6}{HTML}{97A7BD}%

\definecolor{G1}{HTML}{EDF8F4}%
\definecolor{G2}{HTML}{D8EDE8}%
\definecolor{G3}{HTML}{C4DED7}%
\definecolor{G4}{HTML}{B3CFC7}%
\definecolor{G5}{HTML}{A3C1BA}%
\definecolor{G6}{HTML}{94B3AC}%

\definecolor{N1}{HTML}{F5F3F0}%
\definecolor{N2}{HTML}{E9E5E1}%
\definecolor{N3}{HTML}{DDD9D5}%
\definecolor{N4}{HTML}{D1CCC7}%
\definecolor{N5}{HTML}{C5BFB9}%
\definecolor{N6}{HTML}{BAB3AD}%

\definecolor{R7}{HTML}{FDEEEE}%
\definecolor{R8}{HTML}{FCE9E7}%
\definecolor{R9}{HTML}{F9E4E2}%
\definecolor{R10}{HTML}{F7DDDB}%
\definecolor{R11}{HTML}{F5D7D4}%
\definecolor{R12}{HTML}{F3D2CF}%

\definecolor{R13}{HTML}{F1C8C4}%
\definecolor{R14}{HTML}{ECC0BC}%
\definecolor{R15}{HTML}{E8B9B5}%
\definecolor{R16}{HTML}{E3B1AE}%
\definecolor{R17}{HTML}{DFA8A4}%
\definecolor{R18}{HTML}{DA9F9A}%

\definecolor{R19}{HTML}{D49792}%
\definecolor{R20}{HTML}{CF8E8A}%
\definecolor{R21}{HTML}{C98583}%
\definecolor{R22}{HTML}{C27B7A}%
\definecolor{R23}{HTML}{BC7171}%
\definecolor{R24}{HTML}{B66768}%

\definecolor{O1}{HTML}{FFF4E8}%
\definecolor{O2}{HTML}{FDEFE0}%
\definecolor{O3}{HTML}{FBE9D8}%
\definecolor{O4}{HTML}{F8E3CF}%
\definecolor{O5}{HTML}{F5DCC6}%
\definecolor{O6}{HTML}{F2D6BE}%

\definecolor{O7}{HTML}{F0CDB1}%
\definecolor{O8}{HTML}{EDC5A7}%
\definecolor{O9}{HTML}{EABD9D}%
\definecolor{O10}{HTML}{E7B593}%
\definecolor{O11}{HTML}{E4AD89}%
\definecolor{O12}{HTML}{E0A580}%

\definecolor{O13}{HTML}{D89B75}%
\definecolor{O14}{HTML}{D1926B}%
\definecolor{O15}{HTML}{C98762}%
\definecolor{O16}{HTML}{C27D59}%
\definecolor{O17}{HTML}{BA734F}%
\definecolor{O18}{HTML}{B26946}%

\definecolor{PoppyOrange1}{HTML}{FF8C3A}%
\definecolor{PoppyOrange2}{HTML}{FF7A17}%
\definecolor{PoppyOrange3}{HTML}{FF6A00}%
\definecolor{PoppyOrange4}{HTML}{E65D00}%
\definecolor{PoppyOrange5}{HTML}{D45700}%
\definecolor{PoppyOrange6}{HTML}{FF9F54}%
\definecolor{PoppyOrange7}{HTML}{FFB378}%

\definecolor{MutedRose}{HTML}{E7A4A4}%

\definecolor{ActionRed}{HTML}{D45A5A}%
\definecolor{SignalRed}{HTML}{C63B3B}%
\definecolor{DeepProcessRed}{HTML}{B62A2A}%

\definecolor{LightRoseTint}{HTML}{F0BBBB}%
\definecolor{WarmCoral}{HTML}{E08F8F}%
\definecolor{ProcessShade}{HTML}{AF4444}%
\definecolor{DarkBurgundy}{HTML}{8F2F2F}%

\definecolor{tabhl}{HTML}{E3EEF8}%

\def\ourmethod{LaGS}
\title{Latent Gaussian Splatting for 4D Panoptic Occupancy Tracking}

\author{Maximilian Luz$^{1}$, Rohit Mohan$^{1}$, Thomas Nürnberg$^{2}$, Yakov Miron$^{2,3}$, Daniele Cattaneo$^{1}$, and Abhinav Valada$^{1}$%
\thanks{
Manuscript received 21 December 2025; accepted 13 May 2026. This article was recommended for publication by Associate Editor G. Dubbelman and Editor M. Vincze upon evaluation of the reviewers’ comments.}
\thanks{The work of Abhinav Valada was supported in part by the Deutsche Forschungsgemeinschaft (DFG, German Research Foundation) under Grant 539134284, in part by EFRE under Grant FEIH\_2698644, and in part by the State of Baden-Württemberg. This work was supported by the Bosch Research Collaboration on AI-Driven Automated Driving.
}%
\thanks{$^{1}$Department of Computer Science, University of Freiburg, Germany.}%
\thanks{$^{2}$Bosch Research, Robert Bosch GmbH, Germany.}%
\thanks{$^{3}$Leon H.\ Charney School of Marine Sciences, University of Haifa, Israel.}%
\thanks{Corresponding author: \href{mailto:<luz@cs.uni-freiburg.de>}{luz@cs.uni-freiburg.de}}
\thanks{
\copyright 2026 IEEE. This article has been accepted for publication in IEEE Robotics and Automation Letters (RA-L). The final authenticated version is available online at \url{https://doi.org/10.1109/LRA.2026.3703990}.
}
}

\begin{document}
\maketitle

\begin{abstract}
Capturing 4D spatiotemporal scene structure is crucial for the safe and reliable operation of robots in dynamic environments.
However, existing approaches typically address only part of the problem: they either provide coarse geometric tracking via bounding boxes or detailed 3D occupancy estimates that lack explicit temporal association and instance-level reasoning.
In this work, we present Latent Gaussian Splatting ({\ourmethod}) for 4D Panoptic Occupancy Tracking (4D-POT).
We revisit the underlying representation and model 3D features as a sparse set of feature-bearing Gaussians.
These act as dynamic, volume-oriented keypoints that enable spatially continuous, distance-weighted aggregation of multi-view features before being splatted into a voxel grid for decoding.
This point-centric formulation enables flexible, data-dependent receptive fields and long-range spatial interactions that are difficult to capture with local and dense voxel-based operators.
A hierarchical Gaussian representation further enables multi-scale reasoning by combining global context from coarse super-points with fine-grained detail from higher-resolution streams.
Extensive experiments on Occ3D nuScenes and Waymo demonstrate state-of-the-art performance for 4D-POT.
We provide code and models at \url{https://lags.cs.uni-freiburg.de/}.
\end{abstract}


\section{Introduction}
\label{sec:intro}

\IEEEPARstart{C}{amera}-based 4D panoptic occupancy tracking (4D-POT)~\cite{chen2025trackocc} combines dense geometric reconstruction, semantic understanding, and temporal consistency within a unified framework.
This holistic formulation promises a comprehensive representation of dynamic environments, addressing key shortcomings of prior paradigms \cite{luz2024AmodalOpticalFlow,mohan2026upfuse}.
Classical box-based tracking methods model scenes using coarse cuboids~\cite{buchner20223d,lang2023self,kappeler2025bridging}, lacking fine-grained geometry and volumetric semantics, while standard 3D occupancy prediction~\cite{zhang2023OccFormer,wei2023SurroundOcc,ma2024COTR} typically operates per frame, producing dense voxel grids without instance identities or temporal association.
In contrast, 4D-POT assigns a semantic class to every voxel while simultaneously distinguishing and tracking individual object instances across time, bridging geometric fidelity and instance-level awareness.

\begin{figure}
    \centering
    \input{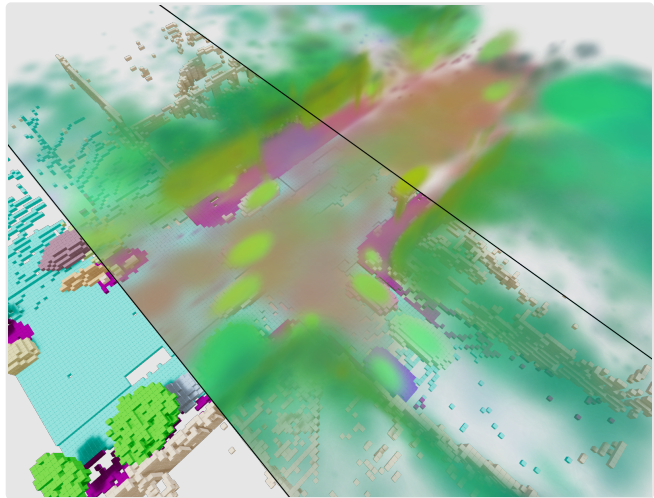}
    \caption{%
        Illustration of our latent Gaussian representation.
        Bottom left: panoptic voxel predictions.
        Center: latent Gaussian features colorized via principal component analysis, splatted to 2D, and overlaid over voxel predictions.
        Top right: latent Gaussians.%
    }
    \label{fig:teaser}
\end{figure}

Despite the relative novelty of 4D-POT, existing approaches largely follow a direct composition of mask-based 3D occupancy prediction with query-based end-to-end 3D multi-object tracking (3D MOT)~\cite{chen2025trackocc}, typically relying on dense voxel feature encoders adapted from 3D occupancy prediction.
While effective, these designs inherit limitations of dense voxel representations, including restricted receptive fields and limited flexibility in modeling long-range spatial interactions.

In this work, we revisit the underlying representation and propose to model 3D features as a sparse set of feature-bearing Gaussians.
Concretely, we treat 3D Gaussians as dynamic, volume-oriented keypoints that carry feature embeddings and spatial extent, enabling spatially continuous, distance-weighted aggregation of sparse features into dense voxel representations via Gaussian splatting.
This yields a point-centric formulation in which features are aggregated in a sparse latent space before being splatted back into a voxel grid for downstream processing.
We term this concept \emph{Latent Gaussian Splatting} ({\ourmethod}), illustrated in \cref{fig:teaser}.
By replacing dense voxel feature volumes with a sparse Gaussian representation, {\ourmethod} enables more flexible, data-dependent receptive fields and long-range spatial interactions that are difficult to capture with local voxel-based operators such as COTR~\cite{ma2024COTR}.
Furthermore, by leveraging hierarchical Gaussian representations, we construct coarse super-points that support global reasoning while preserving fine-grained detail in higher-resolution point sets.
This hierarchical formulation enables multi-scale reasoning, where coarse representations capture global context while fine streams preserve local detail, providing a principled balance between local detail and global context within a unified encoder design.%

Beyond this representation, we identify an additional challenge in mask-based 4D-POT:
the increased number of instance queries introduces substantial memory demands, as query-based tracking typically requires many more queries than the number of expected objects~\cite{meinhardt2022TrackFormer,zeng2022MOTR,zhang2022MUTR3D,pang2023PFTrack}.
This is further exacerbated when backpropagating gradients across multiple frames, as in prior end-to-end tracking approaches~\cite{zhang2022MUTR3D,pang2023PFTrack}.
However, we observe that propagating detached queries and optimizing each frame independently maintains strong performance while significantly reducing resource requirements and allowing capacity to be reallocated to the decoder transformer.%

In summary, our primary contributions are five-fold:
\textbf{(1)} We introduce \emph{Latent Gaussian Splatting}, a sparse, feature-bearing Gaussian representation for dense 3D/4D prediction that enables spatially continuous aggregation and more flexible, expressive feature encoding.
\textbf{(2)} We design a hierarchical Gaussian encoder that enables multi-scale reasoning by combining coarse global context with fine-grained local detail within a unified point-centric formulation.
\textbf{(3)} We re-evaluate existing 4D-POT metrics, address inconsistencies in prior implementations, and extend evaluation to nuScenes~\cite{caesar2020nuscenes} with newly generated ground-truth annotations and reproduced baselines.
\textbf{(4)} We achieve state-of-the-art performance on Occ3D nuScenes and Waymo, improving by up to \SI[explicit-sign=+]{15.2}{\pp} STQ on nuScenes.
\textbf{(5)} We provide code and models at \url{https://lags.cs.uni-freiburg.de/}.

\section{Related Work}
\label{sec:related_work}
In this section, we first briefly review methods for 3D occupancy prediction, followed by a discussion of sparse representations addressing the dense 3D nature of the task, and finally cover 4D panoptic occupancy tracking.

{\parskip=2pt
\textit{3D Occupancy Prediction}: 
While the majority of 3D perception in autonomous driving still relies on bounding-box representations, the promise of finer geometric details has led to significant advances in 3D semantic occupancy prediction in recent years~\cite{ma2024COTR, mohan2026forecastocc}.
Notably, Occ3D~\cite{tian2023occ3d} extends the widely adopted nuScenes~\cite{caesar2020nuscenes} and Waymo~\cite{sun2020waymo} datasets to semantic occupancy prediction, providing more challenging dynamic scenes.
Current state-of-the-art thereon \cite{zhang2023OccFormer,wei2023SurroundOcc,ma2024COTR} largely follows MaskFormer~\cite{cheng2021MaskFormer}, employing a mask-based transformer decoder and posing the task as 3D segmentation.
SparseOcc~\cite{tang2024SparseOcc} and PaSCo~\cite{cao2024PaSCo} show that, akin to 2D panoptic image segmentation, such a decoder can be adapted straightforwardly to predict 3D panoptic occupancy.
}

{\parskip=2pt
\textit{Sparse 3D Occupancy Prediction}: 
A major practical challenge in 3D semantic occupancy prediction is the dense 3D nature of the task.
Several works explore compressed representations~\cite{huang2023TPVFormer,ma2024COTR} or exploit the intrinsic sparsity of the task by focusing only on occupied regions~\cite{tang2024SparseOcc,wang2024OPUS}, yet remain largely retain voxel-aligned.
Notably, GaussianFormer~\cite{huang2025GaussianFormer} proposes 3D Gaussians for representing both occupied and free space, followed by GaussianFormer-2~\cite{huang2025GaussianFormer2}, only modeling occupied space for a sparse object-centric approach.
Both methods splat the predicted 3D Gaussians to the final voxel grid for the task output.
We argue this creates an opportunity to integrate ideas from recent point-based 3D perception methods~\cite{liu2024LION,wu2024PointTransformerV3}.
Specifically, Gaussians can be treated as superpoints, yielding a sparse latent representation that alleviates costly dense 3D processing while remaining convertible to dense voxel representations via splatting.
Our novelty, therefore, lies in treating splatting as an intermediate step in the encoder, splatting features instead of appearance or semantics.
Moreover, the sparse representation enables more effective reasoning across larger and more flexible neighborhoods, improving information exchange, 2D-to-3D lifting, and scalability.
While sparsity has also been explored in classical object detection~\cite{chen2023VoxelNeXt}, the prevalent architectures remain dense.
}

{\parskip=2pt
\textit{4D Panoptic Occupancy Tracking}:
By extending 3D panoptic occupancy prediction temporally, TrackOcc~\cite{chen2025trackocc} introduces the task of 4D panoptic occupancy tracking (4D-POT).
While it can be understood as a 3D extension to video panoptic segmentation (VPS)~\cite{kim2020vps}, its dense 3D nature requires specifically tailored approaches due to its inherent computational complexity.
To this end, TrackOcc incorporates ideas from end-to-end 3D multi-object tracking~\cite{meinhardt2022TrackFormer,zeng2022MOTR,zhang2022MUTR3D,pang2023PFTrack}, ensuring temporally consistent instance assignments by propagating instance-mask-queries across frames.
While TrackOcc relies on MUTR3D~\cite{zhang2022MUTR3D} for query-based tracking, we instead build on PF-Track~\cite{pang2023PFTrack}, leveraging its lightweight temporal query refinement, and further streamline our approach to allow for more than one decoder layer, significantly improving tracking performance.
}

\section{Method}
\label{sec:method}

\begin{figure*}%
    \centering
    \begin{adjustbox}{center,totalheight=0.275\textheight}%
        \input{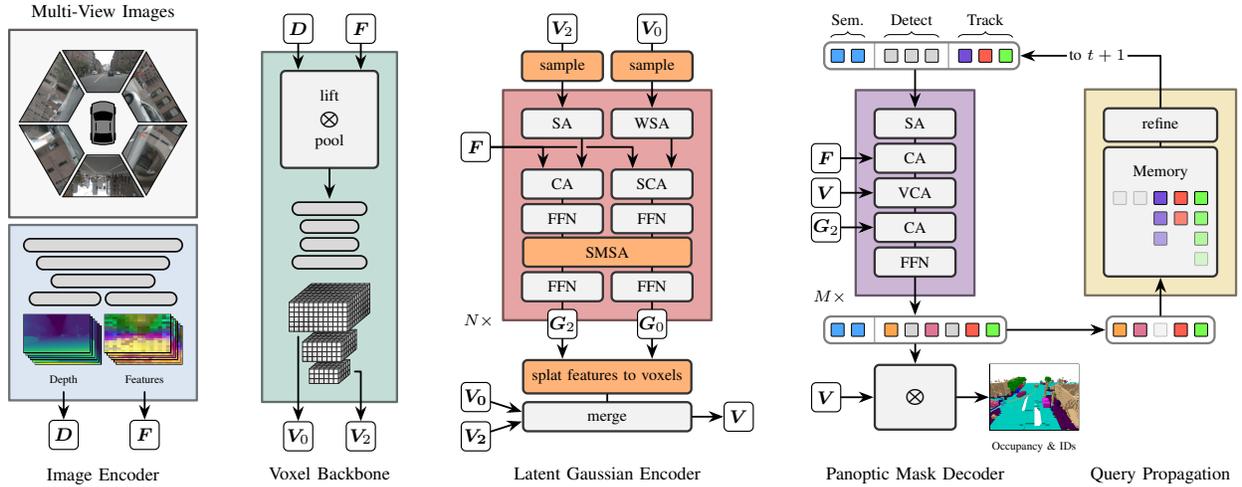}
    \end{adjustbox}%
    \caption[Architecture overview.]{
        Architecture overview (left to right).
        An image encoder (\colorboxmark{B2}) produces image features $\tensor{F}$ and depth $\tensor{D}$.
        Features are lifted via depth to a 3D volume and further encoded into a 3D voxel feature pyramid ($\tensor{V}_0$, $\tensor{V}_2$, \colorboxmark{G3}).
        Our latent Gaussian encoder (\colorboxmark{MutedRose}) samples points from the pyramid volumes and processes them in a coarse (left) and fine (right) stream, using self-attention (SA), windowed self-attention (WSA), cross-attention (CA), spatial cross-attention (SCA), and feed-forward networks (FFN).
        Our novel Serialized Multi-Stream Attention (SMSA) facilitates information exchange between streams.
        Refined points are decoded as Gaussians ($\tensor{G}_0$, $\tensor{G}_2$) and splatted back to a 3D feature volume, which is then refined to the final voxel volume $\tensor{V}$.
        Our transformer decoder (\colorboxmark{P4}) then decodes this volume into semantic and instance masks using volume cross-attention (VCA) for efficient query-to-3D-volume attention.
        Tracking is facilitated by the tracking-by-attention paradigm.
        We refine track queries by spatio-temporal reasoning before passing them onto the next frame (\colorboxmark{Y2}).
    }
    \label{fig:architecture}
\end{figure*}

4D panoptic occupancy tracking requires the tight integration of geometric reconstruction, semantic understanding, and temporal association (\cref{ssec:task_definition}).
Our approach (Fig.~\ref{fig:architecture}) combines mask-based 3D occupancy prediction~\cite{chen2025trackocc,ma2024COTR} with a query-based 3D multi-object tracking decoder following the tracking-by-attention paradigm~\cite{pang2023PFTrack}.
We largely adopt existing designs for image encoding, lifting, and voxel-based feature extraction, and build upon a PETR-style tracking decoder~\cite{pang2023PFTrack,liu2022petr}.
Our core novelty lies in the representation and 3D encoder design.
Specifically, we introduce a latent Gaussian feature representation, replacing dense voxel processing with a hierarchical point-based representation, and propose Serialized Multi-Stream Attention (SMSA) for efficient interaction across hierarchical point sets.
We further extend Gaussian splatting to feature aggregation and adapt the decoder to jointly operate on voxel and Gaussian features.

Our pipeline first lifts multi-view image features into voxels (\cref{ssec:lifting}), refines them via the latent Gaussian encoder (\cref{ssec:encoder}), and decodes them into semantic and instance masks (\cref{ssec:decoder}), with tracking performed via propagated queries (\cref{ssec:tracking}).
Further details on training and inference are provided in \cref{ssec:training,ssec:inference}.

\subsection{Task Definition}
\label{ssec:task_definition}

Given a consecutive sequence of multi-view images $\tensor{I} = \{I^j_t, I^j_{t-1}, \ldots I^j_{t-T}\}_{j=1}^N$, where $t$ is the current timestep, $T$ the sequence length, and $N$ the number of camera views, camera-based 4D panoptic occupancy tracking requires joint prediction of occupancy, semantics, and instance associations for each 3D voxel surrounding the ego-vehicle \cite{chen2025trackocc}.
Specifically, the task requires predicting $(c_{p,t}, i_{p,t})$ for each voxel position $p$, where $c_{p,t}$ represents the semantic class and occupancy state (i.e., category including \emph{unknown/other} and \emph{free}) and $i_{p,t}$ the associated instance ID at timestep $t$.
Notably, semantic predictions incorporate both tracked \emph{thing} and non-tracked \emph{stuff} classes, while instance IDs are only assigned for tracked \emph{thing} classes.
Intrinsic and extrinsic parameters of all cameras as well as ego-motion of the vehicle are assumed to be known.\looseness=-1

\subsection{Image Encoder, Explicit Lifting, and Voxel Backbone}
\label{ssec:lifting}

Following prior work~\cite{ma2024COTR,chen2025trackocc}, we begin by extracting multi-view image features $\tensor{F}$ with an off-the-shelf image encoder (\cref{fig:architecture}, \colorboxmark{B2}).
A small head is trained to predict a binned depth distribution $\tensor{D}$ independently for each image, which is then used to lift the image features (explicitly) to 3D via the outer product $\tensor{P} = \tensor{F} \otimes \tensor{D}$, creating a pseudo point cloud.
This is subsequently pooled to a 3D voxel grid representation and further refined into a voxel feature pyramid ($\tensor{V}_0$ at output resolution, $\tensor{V}_2$ at $\nicefrac{1}{4}$ output resolution) using standard 3D convolutions~\cite{chen2025trackocc,ma2024COTR} (\cref{fig:architecture}, \colorboxmark{G3}).

\subsection{Latent Gaussian Occupancy Feature Encoder}
\label{ssec:encoder}

Instead of further refining the dense voxel representation, as done in COTR~\cite{ma2024COTR}, we introduce a representation of 3D features as a set of latent Gaussians, treating them as volumetric keypoints.
This point-centric formulation allows us to leverage insights from recent point transformer architectures~\cite{wu2024PointTransformerV3}.
By serializing points via space-filling curves, we obtain larger and more flexible receptive fields than dense 3D operations (e.g., voxel self-attention in COTR~\cite{ma2024COTR}), improving scalability and replacing fixed neighborhoods with learned relations via positional encoding.
Concretely, we represent latent Gaussian queries as feature-coordinate pairs $\tensor{G}_s = \{(\tensor{q}_{s,i}, \tensor{c}_{s,i})\}_{i=1}^{k_s}$, where $\tensor{q}_{s,i} \in \mathbb{R}^C$ denotes the feature embedding and $\tensor{c}_{s,i} \in \mathbb{R}^3$ its 3D coordinate.
The coordinates are inherited from voxel locations during the sampling process described below.
Our encoder (\cref{fig:architecture}, \colorboxmark{MutedRose}) follows a hierarchical design with two streams: a high-resolution (fine) stream $\tensor{G}_0$ with more points for capturing local detail, and a coarse stream $\tensor{G}_2$ with fewer points for efficient global reasoning.

{\parskip 2pt plus2pt
\textit{Voxel-to-Gaussian Sampling}:
We convert voxel features into point representations via a four-step process: (1) compute voxel-wise priority scores, (2) suppress spatial redundancy via max-pooling, (3) sample indices using multinomial sampling, and (4) map sampled indices back to the original voxel grid.
Given a voxel feature volume $\tensor{V}_s$, we use feature magnitudes as scalar priorities and apply 3D max-pooling (kernel size $n=2$) to reduce spatial redundancy, yielding pooled priorities and an argmax index map.
We normalize the pooled priorities to obtain a probability distribution and sample $k_s$ indices without replacement using multinomial sampling, which are mapped back to the original resolution via the argmax map.
The resulting voxel features $\tensor{q}_{s,i}$ and their spatial coordinates $\tensor{c}_{s,i}$ define the initial point set $\tensor{G}_s$, enabling the transition from dense voxel grids to sparse point-based representations.
}

{\parskip 2pt plus2pt
\textit{Point-Based Refinement (Fine Stream)}:
Following Point Transformer V3~\cite{wu2024PointTransformerV3}, we serialize points via a space-filling curve applied to the coordinates $\tensor{c}_{0,i}$ and process the resulting linearized sequence with a multi-layer transformer.
We employ sliding-window self-attention (WSA) with overlapping windows of size $w$, centered at each position $i$, such that attention is restricted to local neighborhoods via masking.
This ensures that memory scales linearly with the number of points $k_0$, while maintaining effective local context.
To efficiently integrate image features, we use spatial cross-attention (SCA)~\cite{li2025BEVFormer}, projecting points into image space and applying 2D deformable attention.
}

{\parskip 2pt plus2pt
\textit{Hierarchical Extension (Coarse Stream)}: 
We extend the encoder by introducing hierarchical streams derived from a 3D feature pyramid ($\tensor{V}_0$, $\tensor{V}_2$).
Instead of sampling from a single resolution, we construct streams $\tensor{G}_0$ and $\tensor{G}_2$ independently from different scales and process them independently through their respective transformer blocks.
While the fine stream ($\tensor{G}_0$) employs sliding-window self-attention together with spatial cross-attention, enabling efficient local refinement, the coarse stream ($\tensor{G}_2$) uses full self-attention (SA) and full image-point cross-attention (CA), leveraging its lower point count to capture more global context.

\parskip 0pt
Interaction between streams is performed exclusively via a novel Serialized Multi-Stream Attention (SMSA) module, enabling controlled information exchange between resolutions while preserving independent per-stream processing.
SMSA merges all streams and, similar to the single-stream case, re-serializes the combined set using a space-filling curve, producing a unified linear sequence.
Sliding-window self-attention (WSA) is then applied to this sequence, enabling local interactions and cross-stream information exchange as points from different streams are interleaved based on spatial proximity.
Positional encoding is applied via Fourier feature embeddings of the coordinates and added to the input keys and queries prior to attention.
After attention, features are restored to the original ordering using the inverse serialization permutation and split back into the respective streams.
SMSA naturally handles varying point densities across streams without requiring explicit balancing.
}

{\parskip 2pt plus2pt
\textit{Gaussian Feature Aggregation}: 
Our downstream architecture depends on a 3D feature volume.
To this end, we extend 3D Gaussian occupancy splatting \cite{huang2025GaussianFormer2} from occupancy prediction to feature aggregation.
For each latent query $\tensor{q}_i$, we predict centers $\tensor{\mu}_i = \tensor{c}_i + \tensor{\delta}_i$ (with offsets $\tensor{\delta}_i$), covariances $\Sigma_i$ (from scale and rotation), opacities $\alpha_i$, and feature embeddings $\tensor{e}_i \in \mathbb{R}^C$.
From these, occupancy $o$ and voxel features $\tensor{f}$ are computed as
\begin{align}
    o(\tensor{x})
        &= 1 - \prod_i \Big(
            1 - \exp\big( -\tfrac{1}{2}\|\tensor{x}-\tensor{\mu}_i\|^2_{\Sigma_i^{-1}} \big)
        \Big), \\
    \tensor{f}(\tensor{x})
        &= o(\tensor{x}) \cdot \frac
            {\sum_i \alpha_i \mathcal{G}_i(\tensor{x}) \tensor{e}_i}
            {\sum_i \alpha_i \mathcal{G}_i(\tensor{x})},
\end{align}
where $\mathcal{G}_i(\tensor{x}) = \mathcal{N}(\tensor{x} \mid \tensor{\mu}_i, \Sigma_i)$ is the 3D Gaussian PDF, and $\|\tensor{v}\|_{M}^2 = \tensor{v}^\intercal M \tensor{v}$ denotes Mahalanobis distance.
Note that the decoded Gaussians only represent occupied space.
Hierarchical streams are concatenated before aggregation.
Subsequently, we merge the aggregated feature volume with the initial feature pyramid volumes, essentially introducing skip connections and creating a U-Net-like structure.}

\subsection{Panoptic Mask Decoder}
\label{ssec:decoder}

We build upon the PETR-style mask transformer decoder (\cref{fig:architecture}, \colorboxmark{P4}) used in PF-Track~\cite{pang2023PFTrack}, employing detection queries for instance (\emph{thing}) segmentation and semantic queries for global instance-less (\emph{stuff}) masks.
We retain the query-based transformer formulation, but extend it with additional cross-attention to 3D voxel features via deformable attention and, for hierarchical Gaussian encoding, to refined coarse Gaussian features.
Furthermore, we introduce a voxel mask prediction head for volumetric decoding.
After refinement by the transformer, queries are decoded into a mask embedding and semantic class scores.
Similar to MaskFormer~\cite{cheng2021MaskFormer}, binary occupancy masks are computed via a dot product between mask embeddings and voxel features.

\subsection{Tracking, Query Propagation, and Refinement}
\label{ssec:tracking}

To facilitate tracking, we propagate decoded detection queries across frames following the tracking-by-attention paradigm~\cite{meinhardt2022TrackFormer,zeng2022MOTR}.
As in TrackOcc~\cite{chen2025trackocc}, we separate detection (\emph{thing}) and semantic (\emph{stuff}) queries, propagating only detection queries temporally while re-initializing semantic queries for each frame.
The propagated queries, together with newly initialized detection and semantic queries, form the decoder input for the subsequent frame.
New detections initialize tracks while propagated queries maintain existing ones.
To further improve tracking performance, we adopt the spatio-temporal refinement module of PF-Track~\cite{pang2023PFTrack} (\cref{fig:architecture}, \colorboxmark{Y2}) without modification, refining queries based on memory (past) and trajectory prediction (future).
Predicted trajectories are used to fill gaps for intermittently missed or low-confidence detections.%

\subsection{Training and Supervision}
\label{ssec:training}

We supervise our approach on multiple levels. 
Following COTR~\cite{ma2024COTR}, we supervise the depth prediction for explicit feature lifting in the encoder via sparse depth from LiDAR.
Further, we add a small head to decode semantic scores for each Gaussian, splatting them to a semantic voxel grid for direct supervision via a cross-entropy loss, akin to GaussianFormer-2~\cite{huang2025GaussianFormer2}.
The decoder is supervised via both semantic and instance masks as well as box predictions (akin to center supervision in TrackOcc~\cite{chen2025trackocc}).
For detection queries, we use bipartite matching, considering both predicted boxes and masks to assign ground truth instances.
Once a ground-truth instance has been assigned, it is kept across all subsequent training frames.
For semantic queries, we use bipartite matching with masks only.
Supervision of the spatio-temporal refinement module follows PF-Track~\cite{pang2023PFTrack}.

To enable temporal supervision, we train on short multi-frame sequences.
We prevent linear scaling of gradient buffers by detaching track and detection queries after decoding and before refinement, meaning gradients from subsequent frames still flow back to the refinement stage, but not the decoder itself, decoupling individual frames.

\subsection{Inference}
\label{ssec:inference}

The raw predictions of our approach are class scores $\tensor{c}_q$ and occupancy mask scores $m_{q,\tensor{x}}$ for both instance and \emph{stuff} queries.
We score queries via the maximum class score, i.e., $s_q = \|\tensor{c}_q\|_\infty$, filter out inactive ones via a threshold, and compute the dominant query $\hat{q}_{\tensor{x}} = \operatorname{arg\,max}_{q}\{s_q \cdot m_{q,\tensor{x}}\}$ for each voxel $\tensor{x}$.
In contrast to previous methods \cite{chen2025trackocc,cheng2021MaskFormer,cheng2022Mask2Former}, we compute dominant queries independently for instance and \emph{stuff} classes, merging both afterward by overriding the \emph{stuff} predictions with instance ones.

\section{Experiments}
\label{sec:experiments}

The effectiveness of our proposed approach is demonstrated by extensive evaluation.
To this end, we first describe the datasets used for benchmarking in \cref{ssec:datasets} and the evaluation metric in \cref{ssec:eval_metric}, including necessary revisions to correct existing inaccuracies.
Brief descriptions of the baselines follow in \cref{ssec:baselines}, along with implementation and training details in \cref{ssec:training_protocols}.
Benchmarking results are presented in \cref{ssec:benchmark}, followed by a detailed ablation study of the architectural components (\cref{ssec:ablations}), concluding with qualitative comparisons (\cref{ssec:qualitative}).
All evaluations are performed on the respective validation splits.

\subsection{Datasets}
\label{ssec:datasets}

We evaluate our approach on both nuScenes~\cite{caesar2020nuscenes} and Waymo~\cite{sun2020waymo}, with 3D occupancy ground-truth provided by Occ3D~\cite{tian2023occ3d}.
For both datasets, the spatial range is bounded from \SI{-40}{\meter} to \SI{40}{\meter} for $x$ and $y$ and from \SI{-1}{\meter} to \SI{5.4}{\meter} for $z$, with a voxel size of \SI{0.4}{\meter} in each axis, resulting in a voxel grid resolution of \num{200}$\times$\num{200}$\times$\num{16}.
For Waymo, we follow TrackOcc~\cite{chen2025trackocc} and subsample the dataset at every 5th frame, yielding \num{789} training scenes and \num{202} validation scenes with \num{40} samples each.
For nuScenes, we train and evaluate on the full dataset with \num{700} training and \num{150} validation scenes, with each scene containing around \num{40} samples.

To obtain 4D panoptic occupancy labels for nuScenes, we assign instance IDs to all thing-class voxels of the Occ3D semantic occupancy data by using the ground-truth box labels.
Specifically, instance IDs are assigned based on the intersecting box of the same class.
Ambiguities, such as voxels being intersected by none or multiple boxes, are resolved by choosing the closest instance.
Notably, the assigned instance IDs are consistent with the nuScenes box instance IDs, facilitating direct box-to-voxel correspondence.
We make this data preprocessing available alongside our code.
For Waymo, we rely on the data provided by TrackOcc \cite{chen2025trackocc}.

\subsection{Evaluation Metrics}
\label{ssec:eval_metric}
Following TrackOcc~\cite{chen2025trackocc}, we adapt the Segmentation and Tracking Quality (STQ), originally introduced for video panoptic segmentation \cite{weber2021step}, to 4D occupancy prediction and tracking.
STQ is defined as the geometric mean of Segmentation Quality (SQ) and Association Quality (AQ),
\begin{equation}\label{eq:stq}
    \text{STQ} = \sqrt{\text{SQ} \cdot \text{AQ}},
\end{equation}
where SQ is the classical mean intersection over union (mIoU) of semantic occupancy prediction \cite{behley2019semantickitti,tian2023occ3d}.

AQ represents the mean IoU between each ground-truth and predicted 4D tube, where each individual IoU is weighted by the respective intersected fraction to facilitate a soft assignment and bound the AQ score by one.
Mathematically, we define the 4D panoptic occupancy predictions $P = \{(p, t, i, c)\}$ as the set over tuples of 3D voxel position $p$, time step $t$, instance ID $i$, and class $c$.
Equivalently, we define $G$ as the set of ground-truth tuples.
From this, we derive $P_i = \{ (p, t, i) \mid (p, t, i, c) \in P \}$ and equivalently $G_i$ as the instance prediction and ground-truth for instance $i$.
Finally, we define
\begin{align}\label{eq:aq}
    \text{AQ} &= \frac{1}{|I_G|} \sum_{i \in I_G} \frac{1}{|G_i|} \sum_{j \in I_P} \left| G_i \cap P_j \right| \cdot \frac{|G_i \cap P_j|}{|G_i \cup P_j|},
\end{align}
where $I_G = \{i \mid (p, t, i, c) \in G\}$ is the set of ground-truth instance IDs and $I_P$ is the set of predicted instance IDs.

Extending over TrackOcc, we further propose the \AQs metric for single-frame panoptic assessment.
\AQs is constructed analogously to \cref{eq:aq}, with the exception that we do not consider tracking tubes but only single-frame instances for matching, i.e., enforce $\forall (p_1, t_1, i_1, c_1), (p_2, t_2, i_2, c_2) \in G: t_1 \neq t_2 \Rightarrow i_1 \neq i_2$.
\STQs follows analogously again as the geometric mean over the already non-temporal mIoU and \AQs.
This avoids the well-documented shortcomings of the panoptic quality (PQ) metric \cite{weber2021step}.
In addition to STQ, AQ, and mIoU/SQ, we also use the binary IoU to assess the quality of the binary free/non-free occupancy prediction.

Following Occ3D~\cite{tian2023occ3d}, we evaluate only on visible regions (using the \enquote{camera} mask).
Notably, AQ does not depend on any class assignments, and SQ does not depend on any instance information, strictly separating semantic and instance segmentation between SQ and AQ.
While mathematically sound, however, we find that the metric implementations of TrackOcc~\cite{chen2025trackocc} are flawed: they solely consider areas occupied in the ground-truth data and ignore regions marked as free space.
This skews the metric significantly, as any false positives in known free space are disregarded, essentially only counting true positives and false negatives.
Mathematically, this is equivalent to applying \cref{eq:aq} to $P_i' = P_i \cap M$ and $G_i' = G_i \cap M$, where $M$ is a mask indicating occupied space in $G$.
We therefore reconstruct and re-evaluate the baselines presented by TrackOcc, as well as TrackOcc itself, and provide corrected implementations alongside our code.

\subsection{Baselines}
\label{ssec:baselines}

To ensure fair comparison under the revised metrics, we reimplement the baselines proposed by TrackOcc~\cite{chen2025trackocc} within a unified framework.
As the original baseline implementations are not publicly available, we build upon TrackOcc and adapt it into a single-frame 3D panoptic occupancy model by removing query propagation, which serves as a common backbone for all methods.
On top of this shared setup, we implement different tracking strategies: MinVIS-style bipartite matching of queries across frames based on cosine similarity~\cite{huang2022minvis}, a CTVIS-based variant with learned association via contrastive training~\cite{ying2023ctvis}, heuristic box extraction followed by tracking with AB3DMOT~\cite{weng2020ab3dmot}, and IoU-based matching inspired by 4D-LiDAR panoptic segmentation~\cite{aygun20214dpls}.
For post-processing-based approaches, we tune hyperparameters independently for each dataset.
To evaluate the effectiveness of temporal association, we additionally introduce a Per-Frame baseline, which assigns independent instance IDs at each timestep.
This provides a lower bound for tracking performance, below which temporal association degrades instance segmentation quality.
All baselines share identical voxel encoders, image backbones (ResNet-50), and training protocols.

\begin{table}[t]
    \centering
    \caption{4D-POT performance on Occ3D nuScenes.}
    \label{tab:results_nusc}
    \vspace{-0.75em}
    \footnotesize
\setlength{\tabcolsep}{2.5pt}  
\begin{tabular}{=+l*{8}{+S[table-format=2.1,detect-all]}}
\toprule
    & & & & & \multicolumn{3}{c}{mIoU} & \\
\cmidrule(lr){6-8}
    Approach & {STQ} & {AQ} &  {\STQs} & {\AQs} & {all} & {things} & {stuff} & {IoU} \\
\midrule

Per-Frame &
     9.0 & 2.5 & 21.8 & 14.7 & 32.5 & 26.4 & 41.2 & 63.2
    \\

MinVIS$^{\dagger}$ \cite{huang2022minvis} &
    11.8 &  4.3 & 21.8 & 14.7 & 32.5 & 26.4 & 41.2 & 63.2
    \\

CTVIS$^{\dagger}$ \cite{ying2023ctvis} &
    11.4 &  3.9 & \tabul{22.5} & \tabul{15.4} & \tabul{33.0} & \tabul{27.0} & 41.5 & \tabul{63.8}
    \\

4D-LCA$^{\dagger}$ \cite{aygun20214dpls} &
    12.5 &  4.8 & 21.8 & 14.7 & 32.5 & 26.4 & 41.2 & 63.2
    \\

AB3DMOT$^{\dagger}$ \cite{weng2020ab3dmot} &
    \tabul{13.1} &  \tabul{5.3} & 21.8 & 14.7 & 32.5 & 26.4 & 41.2 & 63.2
    \\

TrackOcc$^{\ddagger}$ \cite{chen2025trackocc} &
    12.2 &  4.7 & 19.7 & 12.1 & 32.1 & 25.3 & \tabul{41.8} & 63.7
    \\

\rowcolor{tabhl}
{\ourmethod}-2s R50 (Ours)
    & \tabbf{27.4}
    & \tabbf{20.9}
    & \tabbf{31.4}
    & \tabbf{27.2}
    & \tabbf{36.2}
    & \tabbf{31.4}
    & \tabbf{43.0}
    & \tabbf{64.4}
    \\

\arrayrulecolor{lightgray}
\midrule
\arrayrulecolor{black}

TrackOcc V99$^{\ddagger}$ \cite{chen2025trackocc}
    & 15.5
    &  6.5
    & 23.8
    & 15.3
    & 37.0
    & 31.0
    & 45.6
    & 67.0
    \\

\rowcolor{tabhl}
{\ourmethod}-2s V99 (Ours)
    & \tabbf{32.3}
    & \tabbf{25.6}
    & \tabbf{36.4}
    & \tabbf{32.5}
    & \tabbf{40.7}
    & \tabbf{37.0}
    & \tabbf{46.0}
    & \tabbf{67.3}
    \\

\bottomrule
\end{tabular}%

\vspace{0.5em}
\begin{minipage}[t]{0.99\linewidth}
    \scriptsize
    $\dagger$: Baselines reproduced by us.
    $\ddagger$: Official code trained for nuScenes/V99 backbone.
\end{minipage}
\end{table}

\begin{table}[t]
    \centering
    \caption{4D-POT performance on Occ3D Waymo.}
    \label{tab:results_waymo}
    \vspace{-0.75em}
    \footnotesize
\setlength{\tabcolsep}{2.5pt}  
\begin{tabular}{=+l*{8}{+S[table-format=2.1,detect-all]}}
\toprule
    & & & & & \multicolumn{3}{c}{mIoU} & \\
\cmidrule(lr){6-8}
    Approach & {STQ} & {AQ} &  {\STQs} & {\AQs} & {all} & {things} & {stuff} & {IoU} \\
\midrule

\grayrow
Per-Frame &
    11.9 &  4.7 & \tabna & \tabna & \tabul{30.0} & \tabul{32.7} & 29.3 & \tabna
    \\

\grayrow
MinVIS$^{\dagger}$ \cite{huang2022minvis} &
    15.0 &  7.5 & \tabna & \tabna & \tabul{30.0} & \tabul{32.7} & 29.3 & \tabna
    \\

\grayrow
CTVIS$^{\dagger}$ \cite{ying2023ctvis} &
    16.4 &  9.3 & \tabna & \tabna & 28.9 & 31.9 & 28.1 & \tabna
    \\

\grayrow
4D-LCA$^{\dagger}$ \cite{aygun20214dpls} &
    16.2 &  8.7 & \tabna & \tabna & \tabul{30.0} & \tabul{32.7} & 29.3 & \tabna
    \\

\grayrow
AB3DMOT$^{\dagger}$ \cite{weng2020ab3dmot} &
    18.0 & 10.8 & \tabna & \tabna & \tabul{30.0} & \tabul{32.7} & 29.3 & \tabna
    \\

\grayrow
TrackOcc$^{\ddagger}$ \cite{chen2025trackocc} &
    \tabul{20.2} & \tabul{13.8} & \tabna & \tabna & 29.4 & 29.7 & \tabul{29.4} & \tabna
    \\

\rowcolor{gray!10}
\grayrow
{\ourmethod}-2s R50 (Ours)
    & \tabbf{26.2}
    & \tabbf{22.0}
    & \tabna
    & \tabna
    & \tabbf{31.1}
    & \tabbf{36.6}
    & \tabbf{29.6}
    & \tabna
    \\

\arrayrulecolor{lightgray}
\midrule
\arrayrulecolor{black}

\grayrow
TrackOcc V99
    & 21.7
    & 14.6
    & \tabna
    & \tabna
    & 32.3
    & 32.5
    & 32.3
    & \tabna
    \\

\rowcolor{gray!10}
\grayrow
{\ourmethod}-2s V99 (Ours)
    & \tabbf{30.1}
    & \tabbf{25.8}
    & \tabna
    & \tabna
    & \tabbf{35.3}
    & \tabbf{41.3}
    & \tabbf{33.6}
    & \tabna
    \\

\midrule
\resetrow
Per-Frame &
     9.1 &  4.0 & 18.1 & 15.6 & 20.9 & 21.9 & 20.6 & 58,4
    \\

MinVIS$^{\dagger}$ \cite{huang2022minvis} &
    11.0 &  5.8 & 18.1 & 15.6 & 20.9 & 21.9 & 20.6 & 58.4
    \\

CTVIS$^{\dagger}$ \cite{ying2023ctvis} &
    12.5 &  7.3 & \tabul{18.7} & \tabul{16.5} & 21.2 & \tabul{22.3} & 20.9 & 59.6
    \\

4D-LCA$^{\dagger}$ \cite{aygun20214dpls} &
    12.1 &  7.0 & 18.1 & 15.6 & 20.9 & 21.9 & 20.6 & 58.4
    \\

AB3DMOT$^{\dagger}$ \cite{weng2020ab3dmot} &
    13.3 &  8.5 & 18.1 & 15.6 & 20.9 & 21.9 & 20.6 & 58.4
    \\

TrackOcc$^{\ddagger}$ \cite{chen2025trackocc} &
    \tabul{15.2} & \tabul{10.7} & 18.1 & 15.2 & \tabul{21.6} & 21.9 & \tabbf{21.5} & \tabbf{60.1}
    \\

\rowcolor{tabhl}
{\ourmethod}-2s R50 (Ours)
    & \tabbf{18.4}
    & \tabbf{15.3}
    & \tabbf{20.6}
    & \tabbf{19.2}
    & \tabbf{22.0}
    & \tabbf{24.3}
    & \tabul{21.4}
    & \tabul{59.8}
    \\

\arrayrulecolor{lightgray}
\midrule
\arrayrulecolor{black}

TrackOcc V99* \cite{chen2025trackocc}
    & 16.7
    & 11.7
    & 20.0
    & 16.7
    & 23.9
    & 24.7
    & \tabbf{23.7}
    & \tabbf{62.9}
    \\

\rowcolor{tabhl}
{\ourmethod}-2s V99 (Ours)
    & \tabbf{21.2}
    & \tabbf{18.3}
    & \tabbf{23.6}
    & \tabbf{22.6}
    & \tabbf{24.6}
    & \tabbf{28.1}
    & \tabbf{23.7}
    & 61.5
    \\

\bottomrule
\end{tabular}%

\vspace{0.5em}
\begin{minipage}[t]{0.99\linewidth}
    \scriptsize
    $\dagger$: Baselines reproduced by us.
    $\ddagger$: Results reproduced with official code and weights.
    *: Official code adapted for and trained with V99 backbone.
    Gray text: metrics as implemented by TrackOcc~\cite{chen2025trackocc}, ignoring false positives.
\end{minipage}
\end{table}

\subsection{Implementation and Training Details}
\label{ssec:training_protocols}

Following PF-Track~\cite{pang2023PFTrack}, we use VoVNetV2-99~\cite{lee2019VoVNetV2} as image backbone with input resolution $800{\times}320$ for nuScenes and $704{\times}256$ for Waymo, and additionally evaluate ResNet-50 for fair comparison.
Image-to-3D lifting follows the BEVDet4D variant of COTR~\cite{ma2024COTR} and TrackOcc~\cite{chen2025trackocc}, with stereo and temporal aggregation disabled due to memory constraints.
All methods are trained on 8 NVIDIA L40 GPUs for 24 epochs with batch size 1 per GPU.
Baselines use a single-frame adaptation of TrackOcc and follow its training procedure.
Our method is trained in two stages (12 epochs pre-training + 12 epochs tracking) using a one-cycle schedule (base LR \num{2e-4}, 500 warmup steps)~\cite{pang2023PFTrack}.
The encoder employs a two-stream latent Gaussian representation with \num{512} coarse and \num{8192} fine points, embedding dimension $C{=}256$, and window size $w{=}1024$, using 3D Hilbert curve serialization.
For VoVNetV2, we use a single coarse image feature scale for spatial cross-attention, whereas for ResNet-50 we follow TrackOcc and use the full feature pyramid.
Both encoder and decoder use 4 transformer layers.
Our model has 76.0M parameters (ResNet-50) and 118.9M parameters (VoVNetV2-99).
Inference takes \SI{374.0}{\milli\second} (FP32) and \SI{352.1}{\milli\second} (BF16), with peak memory usage of \SI{11.2}{\giga\byte} and \SI{5.7}{\giga\byte}, respectively.
We observe no measurable accuracy difference between FP32 and BF16 inference.

\subsection{Comparison with State-of-the-Art}
\label{ssec:benchmark}

\begin{table}
    \centering
    \caption{3D occupancy prediction performance on Occ3D nuScenes.}
    \label{tab:results_occ3d}
    \vspace{-0.75em}
    \footnotesize
\setlength{\tabcolsep}{9pt}  
\begin{tabular}{=+lc*{2}{+S[table-format=2.1,detect-all]}}
\toprule
Approach & Image Backbone
    & {mIoU}
    & {IoU} \\
\midrule

TPVFormer \cite{huang2023TPVFormer}             & ResNet-50  &        34.2  &        66.8  \\
SurroundOcc \cite{wei2023SurroundOcc}           & ResNet-101 &        34.6  &        65.5  \\
OccFormer \cite{zhang2023OccFormer}             & ResNet-50  &        37.4  &        70.1  \\
BEVDet4D \cite{huang2022BEVDet4D}               & ResNet-50  &        39.3  &        73.8  \\
BEVDet4D + COTR \cite{ma2024COTR}               & ResNet-50  & \tabul{44.5} & \tabbf{75.0} \\
BEVDet4D + COTR \cite{ma2024COTR}               & Swin-B     & \tabbf{46.2} & \tabul{74.9} \\

\rowcolor{gray!15}
BEVDet4D + COTR$^{*}$ \cite{ma2024COTR}         & ResNet-50  &        34.6  &        66.3  \\

\midrule

TrackOcc$^{\ddagger}$ \cite{chen2025trackocc}   & ResNet-50  &        32.1  &        63.7  \\

\rowcolor{tabhl}
\resetrow
{\ourmethod-2s} (Ours)                          & VoVNetV2-99 & \tabbf{40.7} & \tabbf{67.3} \\

\bottomrule
\end{tabular}%

\vspace{0.5em}
\begin{minipage}[t]{0.99\linewidth}
    \scriptsize
    Top: 3D semantic occupancy prediction methods without instance segmentation or tracking.
    Bottom: 4D panoptic occupancy tracking methods requiring instance-level modeling and temporal association.
    Gray row/*: Restricted baseline without longterm aggregation and stereo-depth feature lifting, representing the closest comparable setting.
    $\ddagger$: Official code trained for nuScenes.
\end{minipage}
\end{table}

Results for Occ3D nuScenes and Waymo are presented in \cref{tab:results_nusc,tab:results_waymo}.
Across both datasets and backbones, {\ourmethod} consistently outperforms prior approaches in both overall (STQ) and tracking-specific (AQ) metrics.
Using a ResNet-50 (R50) backbone, our method improves over TrackOcc by up to \SI[explicit-sign=+]{15.2}{\pp} STQ and \SI[explicit-sign=+]{16.2}{\pp} AQ on nuScenes, and \SI[explicit-sign=+]{3.4}{\pp} STQ and \SI[explicit-sign=+]{4.6}{\pp} AQ on Waymo, with further gains observed for the stronger VoVNetV2-99 (V99) backbone.
Notably, improvements are most pronounced for instance-related metrics (AQ, \AQs, mIoU-\emph{things}), indicating more effective modeling of object structure and re-identification, while performance on \emph{stuff} classes remains comparable.
The larger gains on nuScenes can be attributed to its more diverse set of \emph{thing} classes, where fine-grained instance discrimination is critical.
In contrast, Waymo contains more diverse \emph{stuff} classes (e.g., bicycle, tree trunk), which may require higher-resolution representations (i.e., more Gaussians) to model adequately.

Further, {\ourmethod} reduces the performance gap between 4D panoptic occupancy tracking and 3D semantic occupancy prediction (mIoU) (\cref{tab:results_occ3d}).
We emphasize that methods in the upper part of the table address only 3D semantic occupancy and do not perform instance segmentation or temporal tracking, resulting in significantly lower memory requirements.
In contrast, 4D-POT requires modeling instance masks and temporal associations, necessitating more queries and higher computational cost.
Consequently, both TrackOcc~\cite{chen2025trackocc} and our method operate under a constrained setting without long-term temporal aggregation or stereo-based lifting, making the gray-shaded BEVDet4D+COTR variant the closest comparable baseline.
The remaining gap to the full BEVDet4D+COTR model can likely largely be attributed to its use of additional components, such as temporal aggregation and multi-frame (\enquote{stereo}) lifting, and the absence of instance-level modeling.

\subsection{Ablation Studies}
\label{ssec:ablations}

\begin{table}[t]
    \centering
    \caption{Ablation study on the latent Gaussian encoder.}
    \label{tab:ablations-encoder}
    \vspace{-0.75em}
    \footnotesize
\setlength{\tabcolsep}{2.4pt}  
\begin{tabular}{=+lcc*{7}{+S[table-format=2.1,detect-all]}}
\toprule
    & & & & & & \multicolumn{3}{c}{mIoU} \\
\cmidrule(lr){7-9}
    Type & Gaussians & Layers & {STQ} & {AQ} & {\AQs} & {all} & {thing} & {stuff} & {IoU} \\
\midrule

\rowcolor{gray!15}
COTR            & \tabna        & 1 &        31.5  & 25.0 & 31.6 & 39.6 & 36.3 & 44.4 & 65.1 \\
{\ourmethod}    & 8192          & 1 &        31.5  & 25.1 & 32.0 & 39.6 & 36.2 & 44.5 & 64.9 \\

\arrayrulecolor{lightgray}
\midrule
\arrayrulecolor{black}

COTR            & \tabna        & 2 &        31.5  & 24.9  & 31.6 & 39.8 & 36.6 & 44.4 & 64.9 \\
{\ourmethod}    & 8192          & 2 &        31.7  & 25.2  & 32.2 & 39.9 & 36.4 & 45.0 & 65.3 \\

\arrayrulecolor{lightgray}
\midrule
\arrayrulecolor{black}

COTR            & \tabna        & 4 &        31.4  &        24.9  &        31.5  &        39.6  &        36.3  &        44.4  &        65.2  \\
{\ourmethod}    & 2048          & 4 & \tabul{32.1} &        25.5  &        32.3  &        40.4  & \tabul{36.8} &        45.5  &        66.3  \\
{\ourmethod}    & 4096          & 4 &        32.0  &        25.4  &        32.2  &        40.3  & \tabul{36.8} &        45.3  &        65.7  \\
{\ourmethod}    & 8192          & 4 &        31.9  &        25.4  & \tabul{32.5} &        40.1  &        36.6  &        44.9  &        65.3  \\

{\ourmethod}-2s & \{512, 2048\} & 4 & \tabbf{32.3} & \tabbf{25.7} & \tabbf{32.7} & \tabul{40.6} & \tabul{36.8} & \tabul{45.9} &        67.0  \\
{\ourmethod}-2s & \{512, 4096\} & 4 & \tabul{32.1} & \tabul{25.6} &        32.4  &        40.4  &        36.5  & \tabul{45.9} & \tabul{67.1} \\

\rowcolor{tabhl}
{\ourmethod}-2s & \{512, 8192\} & 4 & \tabbf{32.3} & \tabul{25.6} & \tabul{32.5} & \tabbf{40.7} & \tabbf{37.0} & \tabbf{46.0} & \tabbf{67.3} \\

\bottomrule
\end{tabular}%

\vspace{0.5em}
\begin{minipage}[t]{0.99\linewidth}
    \scriptsize
    Metrics reported on the nuScenes validation split with VoVNetV2-99 as backbone.
    \tinycolorbox{gray!15}{Gray} indicates the original COTR encoder.
    \tinycolorbox{tabhl}{Blue} indicates our chosen configuration.
\end{minipage}
\end{table}

\begin{table}[t]
    \centering
    \caption{Ablation study on the decoder transformer.}
    \label{tab:ablations-decoder}
    \vspace{-0.75em}
    \footnotesize
\setlength{\tabcolsep}{5pt}  
\begin{tabular}{=+cc*{7}{+S[table-format=2.1,detect-all]}}
\toprule
    & & & & & \multicolumn{3}{c}{mIoU} & \\
\cmidrule(lr){6-8}
    Layers & Refine & {STQ} & {AQ} & \AQs & {all} & {things} & {stuff} & {IoU} \\
\midrule

1 & \cmark &        28.2  &        20.6  &        26.8  & \tabul{38.7} &        34.2  & \tabbf{45.1} &        64.9  \\
4 & \xmark & \tabul{30.5} & \tabul{24.1} & \tabbf{32.4} &        38.5  & \tabul{34.3} & \tabul{44.4} & \tabul{65.0} \\

\rowcolor{tabhl}
4 & \cmark & \tabbf{31.5} & \tabbf{25.0} & \tabul{31.6} & \tabbf{39.6} & \tabbf{36.3} & \tabul{44.4} & \tabbf{65.1} \\

\bottomrule
\end{tabular}%

\vspace{0.5em}
\begin{minipage}[t]{0.955\linewidth}
    \scriptsize
    Performance on the nuScenes validation split reported using the COTR encoder with 1 encoder layer (cf. \cref{tab:ablations-encoder}, first entry).
    \tinycolorbox{tabhl}{Blue} indicates our chosen configuration.
\end{minipage}
\end{table}

{\textit{Latent Gaussian Encoder}:
We evaluate the proposed encoder on the nuScenes validation split (\cref{tab:ablations-encoder}).
With a single transformer layer, our method performs comparably to the COTR~\cite{ma2024COTR} encoder.
As encoder depth increases, however, our approach shows consistent improvements over the voxel-based baseline, particularly for 4-layer configurations, indicating more effective utilization of additional model capacity.
Varying the number of Gaussians leads to non-uniform effects across metrics, reflecting a trade-off between representation granularity and efficiency.
The hierarchical two-stream design further improves performance and yields the best overall results.
This demonstrates that combining fine and coarse representations enhances both local detail and global reasoning, while enabling more effective long-range interactions at coarser resolutions.
We observe approximately linear scaling in runtime with both the number of Gaussians and the encoder depth.%
}%

\begin{figure}[t]
    \centering
    \input{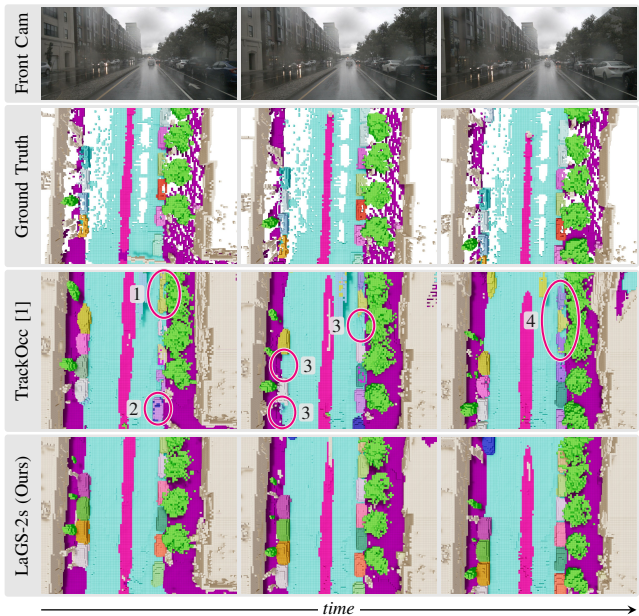}%
    \caption{
        Qualitative results on the Occ3D nuScenes validation split.
        Our approach shows clear improvements in (1) instance separation, (2) instance association, (3) missing detections, and (4) underconfident detections.%
    }
    \label{fig:qualitative_nusc}
\end{figure}

\begin{figure}[t]
    \centering
    \input{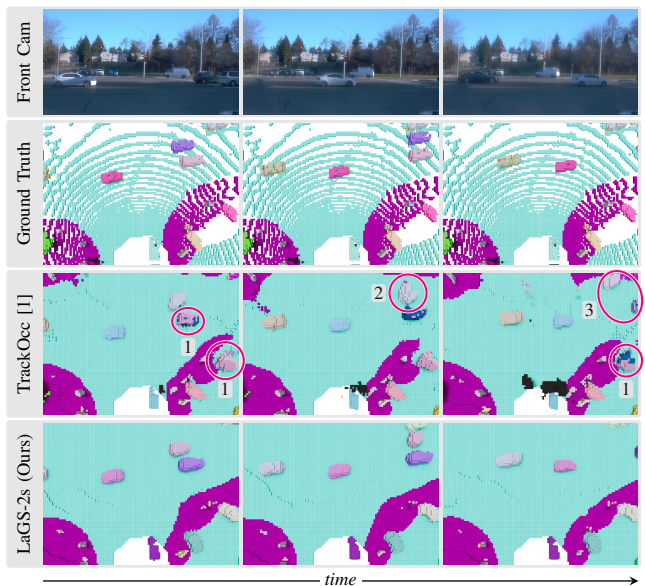}%
    \caption{
        Qualitative results on the Occ3D Waymo validation split.
        Our approach shows clear improvements in (1) instance separation, (2) instance association, and (3) ID switches.%
    }
    \label{fig:qualitative_waymo}
\end{figure}

{\parskip 2pt plus2pt
\textit{Decoder Transformer}:
The decoder plays a central role in instance segmentation and tracking (\cref{tab:ablations-decoder}).
While prior work~\cite{ma2024COTR,chen2025trackocc} typically employs a single decoder layer, increasing the number of layers substantially improves instance segmentation (mIoU-\emph{things}) and tracking quality (AQ), despite the associated increase in computational cost.
This indicates that sufficient decoder capacity is required to fully exploit the improved encoder features.
In addition, spatio-temporal refinement further improves performance, particularly for \emph{thing}-class segmentation, indicating that temporal aggregation enhances both semantic consistency and instance association.
}%

\begin{table}
    \centering
    \caption{Inference runtime and scaling.}
    \label{tab:ablations-scaling}
    \vspace{-0.75em}
    \footnotesize
\setlength{\tabcolsep}{8pt}  
\begin{tabular}{=+lccc+S[table-format=3.1,detect-all]}
\toprule
    \multicolumn{3}{c}{Voxel Encoder} & \multicolumn{1}{c}{Decoder} & \\
\cmidrule(lr){1-3}
\cmidrule(lr){4-4}
    Type & Gaussians & Layers & Layers & {Runtime (ms)} \\
\midrule

COTR            & \tabna        & 1 & 1 & 215.9 \\

\arrayrulecolor{lightgray}\midrule\arrayrulecolor{black}

COTR            & \tabna        & 1 & 4 & 246.3 \\
{\ourmethod}    & 8192          & 1 & 4 & 304.0 \\

\arrayrulecolor{lightgray}\midrule\arrayrulecolor{black}

COTR            & \tabna        & 2 & 4 & 266.8 \\
{\ourmethod}    & 8192          & 2 & 4 & 312.7 \\

\arrayrulecolor{lightgray}\midrule\arrayrulecolor{black}

COTR            & \tabna        & 4 & 4 & 334.4 \\
{\ourmethod}    & 2048          & 4 & 4 & 289.7 \\
{\ourmethod}    & 4096          & 4 & 4 & 307.0 \\
{\ourmethod}    & 8192          & 4 & 4 & 328.3 \\

\rowcolor{tabhl}
{\ourmethod}-2s & \{512, 8192\} & 4 & 4 & 352.1 \\

\bottomrule

\end{tabular}

\vspace{0.5em}
\begin{minipage}[t]{0.99\linewidth}
    \scriptsize
    Runtimes averaged over the Occ3D nuScenes validation split using VoVNetV2-99 as backbone and BF16 precision.
    \tinycolorbox{tabhl}{Blue} indicates our final configuration.
\end{minipage}%
\end{table}

{\parskip 2pt plus2pt
\textit{Runtime and Scaling}:
We analyze the computational cost of our approach in \cref{tab:ablations-scaling}.
Runtime increases predictably with both encoder depth and the number of Gaussians, scaling approximately linearly with depth and moderately with the number of points.
Compared to the voxel-based COTR encoder, our approach incurs additional cost, primarily due to Gaussian-to-voxel feature splatting (up to approx.\ \SI{58}{\milli\second} for \num{8192} Gaussians).
At comparable encoder depth, however, runtimes are similar, and for smaller Gaussian sets (e.g., \num{2048}) our approach can be more efficient, indicating favorable scaling with model capacity.
The hierarchical two-stream variant introduces additional overhead due to cross-stream interaction, but yields the best overall performance.
Increasing the number of decoder layers further increases runtime moderately, yet is crucial for strong instance segmentation and tracking.
Memory usage remains largely constant across configurations (approx.\ \SI{5.7}{\giga\byte} in BF16).
}%

\subsection{Qualitative Evaluation}
\label{ssec:qualitative}

Qualitative results are presented in \cref{fig:qualitative_nusc} (nuScenes) and \cref{fig:qualitative_waymo} (Waymo).
We demonstrate improvements across five categories:
(1) instance separation, correctly separating nearby instances,
(2) instance association, concisely assigning an object to a single instance,
(3) missing detections, leading to incorrect free-space predictions,
(4) underconfident mask predictions, leading to incomplete instance segmentation, and
(5) ID switches, where the same instance ID is wrongly assigned to different objects in subsequent frames.
Across both datasets, {\ourmethod} produces clearer and more consistent instance masks, while TrackOcc~\cite{chen2025trackocc} often exhibits underconfident masks or inconsistent instance assignments (cf.\ highlighted regions).
These observations are consistent with the quantitative instance-level gains reported in \cref{tab:results_nusc,tab:results_waymo}.%

\section{Conclusion}
\label{sec:conclusion}

We proposed a novel Gaussian-driven architecture, outperforming the state-of-the-art in 4D panoptic occupancy tracking.
Inspired by recent advancements in Gaussian splatting and point-transformer methods, we designed a novel 3D occupancy feature encoder. 
Using Gaussians as a 3D representation allows us to convert traditionally dense voxel-grid-based encoders into a sparse, point-wise formulation suitable for transformer-based architectures.
This enables more flexible, data-driven information aggregation, effective 2D-to-3D lifting, and improved scalability in representation capacity.
By utilizing splatting to convert the sparse representation back into a voxel grid, our encoder can replace classical voxel feature encoders while retaining compatibility, albeit at additional computational cost that scales with the number of Gaussians, offering opportunities for further optimization.
Extensive evaluations on both Occ3D nuScenes and Waymo datasets demonstrate the effectiveness of our approach.
We hope that this paves the way for further exploration of dynamic, effective, and saliency-driven 3D representations.

{
    \small
    \bibliographystyle{IEEETran}
    \bibliography{main}
}

\end{document}